%% file: main.tex
\documentclass[11pt]{article} 
\usepackage[]{acl}  
 
\usepackage{times}
\usepackage{latexsym}

\usepackage[T1]{fontenc}

\usepackage[utf8]{inputenc}

\usepackage{microtype}

\usepackage{inconsolata}

\usepackage{graphicx}

\usepackage{caption} 
\usepackage{algorithm}
\usepackage{algorithmic}
\usepackage{xspace}
\usepackage{amssymb}
\usepackage{multirow}

\usepackage{amsmath}
\makeatletter
\@ifpackagelater{lineno}{2023/01/01}{%
  \renewcommand*\linenomathpatch[1]{}%
  \renewcommand*\linenomathpatchAMS[1]{}%
}{}
\makeatother
\usepackage{cleveref}
\crefname{appendix}{appendix}{appendices}
\Crefname{appendix}{Appendix}{Appendices}
\usepackage{newfloat}
\usepackage{listings}
\DeclareCaptionStyle{ruled}{labelfont=normalfont,labelsep=colon,strut=off} 
\floatstyle{ruled}
\newfloat{listing}{tb}{lst}{}
\floatname{listing}{Listing}

\usepackage{booktabs}

\input{macros}

\begin{document}

\title{HybridStitch: Token-Adaptive Model Stitching for Efficient\\Diffusion Generation}
\author{Desen Sun\thanks{These authors contributed equally.} \and
Jason Hon\textsuperscript{\thefootnote} \and
Jintao Zhang \and Sihang Liu}

\maketitle

\input{sec/0_abs}

\input{sec/1_intro}

\input{sec/2_back}
\input{sec/3_method}

\input{sec/4_eval}
\input{sec/5_relate}

\input{sec/6_conclusion}


%
%
\clearpage
\bibliography{ref/ml, ref/sys,ref/misc}
\clearpage
\end{document}

%% file: macros.tex
\newcommand*{\name}{{\texttt{HybridStitch}\xspace}}


%% file: sec/0_abs.tex
\begin{abstract}
  Iterative diffusion generation requires different amounts of model capacity at different token positions. Prior model combinations allocate capacity by denoising step or by drafting and verification role. We present \name{}, a training-free framework that uses token-level feedback to select the model supplying each adopted update. Continuous image refinement and discrete language commitment operate on a shared generation state, while model-specific KV buffers provide attention context for selected queries. \name{} achieves 1.63--1.68$\times$ faster image generation and 1.15--1.58$\times$ diffusion large language model (dLLM) throughput on four main benchmarks. Estimated dLLM decode query cost decreases by 50--66\,\% with comparable generation quality.
\end{abstract}

%% file: sec/1_intro.tex
\section{Introduction}
Diffusion models support high-quality text-to-image generation~\cite{podell2023sdxlimprovinglatentdiffusion,flux2024,labs2025flux1kontextflowmatching,wu2025qwenimagetechnicalreport,esser2024scalingrectifiedflowtransformers,team2025zimage,liu2025decoupled,jiang2025distribution} and increasingly capable language generation~\cite{cheng2025sdarsynergisticdiffusionautoregressionparadigm,nie2025largelanguagediffusionmodels,ye2025dream7bdiffusionlarge}. Both repeatedly evaluate a transformer on an evolving state, making model capacity allocation central to inference cost.

\begin{figure*}[t]
  \begin{center}
  \includegraphics[width=\linewidth]{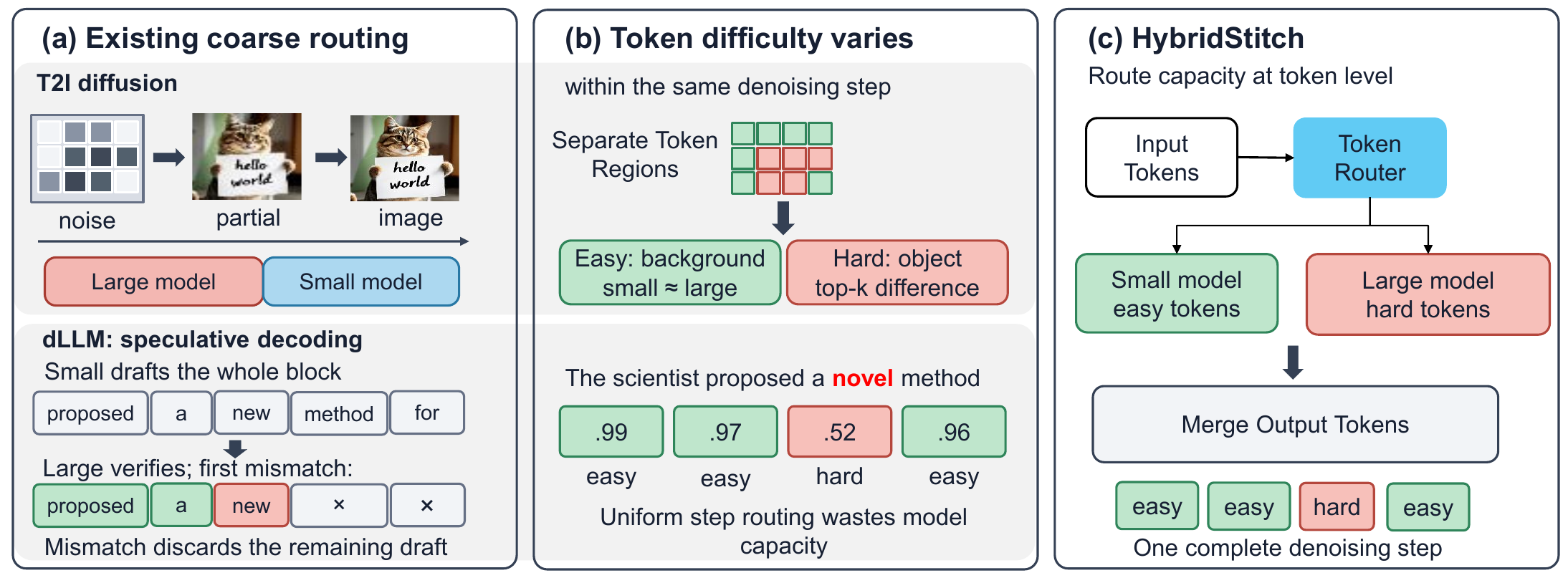}
  \end{center}
  \caption{\label{fig:hybrid_intro} Allocation granularity. (a) Step-level image switching and role-level text speculation. (b) Spatial disagreement and illustrative token confidences. (c) Token-level allocation in \name{}.
  }
\end{figure*}

Model combinations reduce inference cost by assigning part of generation to a smaller model~\cite{cui2026simsdsimplespeculativedecoding,gao2025selfspeculativedecodingdiffusion,pan2025tstitch,cheng2025srdiffusionacceleratevideodiffusion,modm}. T-Stitch and SRDiffusion allocate capacity across denoising steps; speculative decoding assigns drafting and verification roles (\Cref{fig:hybrid_intro}a). Despite the efficiency, these approaches overlook token-level heterogeneity. We observe that, for some tokens, the small model produces outputs nearly identical to those of the large model, whereas other tokens exhibit substantial discrepancies (\Cref{fig:hybrid_intro}b). For example, on MATH500, small-model predictions with confidence above 0.99 match the large model 99.5\,\% of the time, compared with 36.5\,\% below 0.5.

It is intuitive to schedule the model usage at the token-level. However, allocating work at the token level raises two challenges. First, the router must assign ownership before the evaluation that would reveal whether the small model suffices, so it can rely only on feedback from earlier predictions. Second, bidirectional attention lets every selected position depend on the entire state. Therefore, computing a subset of queries requires complete context in each model’s own representation space. An effective scheme must select work from stale but informative feedback and preserve full context under partial execution.

To address these challenges, we present \name{}, a training-free framework that makes each adopted token update the allocation unit (\Cref{fig:hybrid_intro}c). To assign the ownership efficiently, \name{} exploits disagreement and confidence to guide model allocation. \name{} also leverages model-specific KV buffers to supply attention context for selected queries. We apply \name{} to both continuous diffusion, by merging small- and large-model latents, and discrete diffusion, by committing sequences of high-confidence tokens. In either case, subsequent calls receive the shared latent or sequence, allowing each model to use progress made by the other.

The contributions of this paper are as follows: First, we formulate token ownership as a common update object for continuous and discrete diffusion. Second, we realize it through feedback-driven allocation and cache-completed partial execution. Third, we establish the measured quality--efficiency benefits and show that \name{} achieves 1.63--1.68$\times$ image speedup and 1.15--1.58$\times$ throughput on the four main dLLM tasks.

%% file: sec/2_back.tex
\section{Preliminaries and Related Work}

\subsection{Continuous and Discrete Diffusion}
Continuous image generation refines a noisy latent. Our flow-based backends integrate predicted velocities~\cite{esser2024scalingrectifiedflowtransformers}; an Euler update is
\begin{equation}
x_{s+1}=x_s+\Delta\sigma_s\,v_\theta(x_s,\sigma_s),
\end{equation}
where $\sigma_s$ is the noise level, $\Delta\sigma_s=\sigma_{s+1}-\sigma_s$ is the signed step, and $v_\theta$ is the predicted velocity. 

Diffusion large language models (dLLMs), such as LLaDA, Dream, and SDAR, denoise discrete sequences initialized with mask tokens~\cite{nie2025largelanguagediffusionmodels,ye2025dream7bdiffusionlarge,cheng2025sdarsynergisticdiffusionautoregressionparadigm}. Each iteration predicts vocabulary distributions, and a policy selects tokens to unmask. Bidirectional attention conditions predictions on the evolving sequence. In block-causal decoding, an active block attends to itself and the completed prefix. Completed blocks support exact prefix KV reuse; active-block reuse retains previously evaluated representations as the state evolves. Both forms separate prediction from state update: a sampler incorporates continuous predictions, while a commitment policy incorporates discrete predictions.

\subsection{Reusing Computation}
Efficient inference exploits redundancy in iterative generation.
Caching methods reuse intermediate results across denoising steps~\cite{mixfusion,ma2024learningtocache,cache_dit,TaylorSeer} or retrieve states from other requests to skip computation~\cite{modm,sun2024flexcacheflexibleapproximatecache,nirvana}. Attention-sparsity methods reduce attention computation by exploiting locality~\cite{zhangefficient,zhang2026sla,zhang2025spargeattn,zhang2026sla2,xi2025sparse}. For language diffusion, dKV-Cache introduces delayed and conditioned KV reuse, while dLLM-Cache selects response-token updates using feature similarity~\cite{ma2025dkvcache,liu2026dllmcache}. Reuse reduces repeated computation within a model; token-wise capacity allocation additionally selects which model supplies an update. \name{} combines these decisions by executing selected queries against model-specific cached context.

\subsection{Allocating Model Capacity}
Multiple-model inference combines inexpensive prediction with large-model refinement~\cite{modm,pan2025tstitch,cheng2025srdiffusionacceleratevideodiffusion,cui2026simsdsimplespeculativedecoding}.
T-Stitch starts with a small model for early semantic construction and switches to a large model for refinement~\cite{pan2025tstitch}. SRDiffusion starts with a large model and adaptively switches to a small model as generation stabilizes~\cite{cheng2025srdiffusionacceleratevideodiffusion}.
For dLLMs, speculative decoding provides a different way to combine prediction and verification. Self Speculative Decoding uses the dLLM itself as both drafter and verifier, with hierarchical verification of parallel predictions~\cite{gao2025selfspeculativedecodingdiffusion}. SimSD uses draft-model reference tokens and a dedicated attention mask to make multi-token verification valid under changing diffusion contexts~\cite{cui2026simsdsimplespeculativedecoding}. 
The allocation units are denoising steps in T-Stitch and SRDiffusion, drafting and verification roles in speculative decoding, and adopted token updates in \name{}. Generation feedback determines their model ownership within the shared state.

%% file: sec/3_method.tex
\section{Method}

\subsection{Motivation}
\label{subsec:motivation}
\begin{figure}[t]
    \centering
    \includegraphics[width=\linewidth]{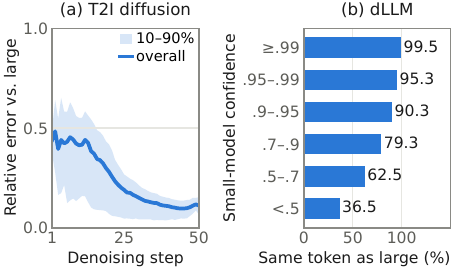}
    \caption{Small--large agreement on identical inputs. (a) SD3 velocity discrepancy and its 10--90\,\% token range at $1024\times1024$. (b) Text agreement by small-model confidence.}
    \label{fig:diff_step}
\end{figure}

Both diffusion tasks, T2I and dLLM, repeatedly predict token updates from an evolving state, and small--large agreement varies across positions. This variation motivates allocating model capacity from local prediction feedback: the small model can handle easier positions cheaply, while harder positions require large-model refinement.

\textbf{T2I.}
Along the sampled trajectory, prediction differences decrease with denoising, while the token-wise range shows spatial variation (\Cref{fig:diff_step}a). Spatial disagreement identifies regions where the two models propose different refinements.

\textbf{dLLM.}
Higher-confidence small-model predictions more often match the large model (\Cref{fig:diff_step}b). Confidence therefore identifies candidates for inexpensive token commitments. In both cases, the generation process supplies the feedback used to allocate subsequent prediction work.

\subsection{Objective}
\begin{figure*}[t]
  \centering
  \includegraphics[width=\linewidth]{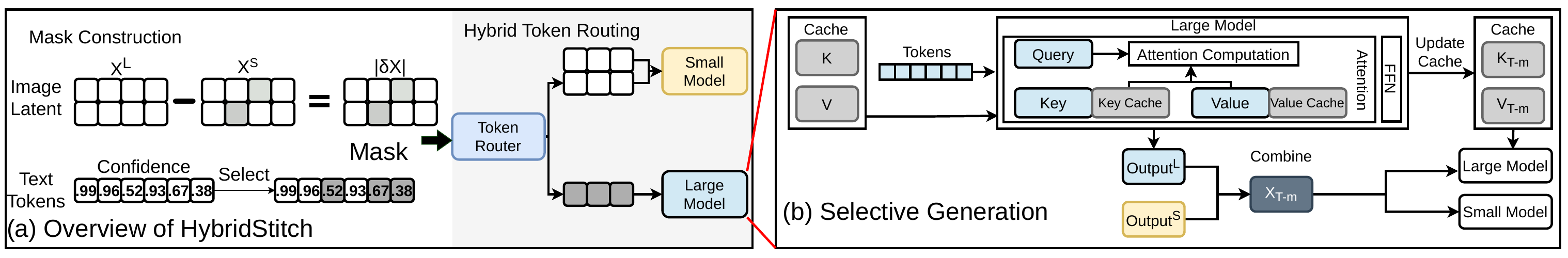}
  \caption{HybridStitch execution framework. (a) Generation feedback produces a routing mask for model-specific token work. (b) Selected queries process attention with cached keys and values.}
  \label{fig:overview}
\end{figure*}

\name{} allocates prediction work between models as the generation state evolves. \emph{Token routing} selects the executing model and prediction positions; \emph{cache completion} supplies full attention context for partial execution (\Cref{fig:overview}). Each iteration routes work, evaluates queries, and updates the shared latent or sequence.

At iteration $s$, let $\pi_s(i)\in\{S,L,\varnothing\}$ identify the owner of position $i$'s adopted update: the small model $S$, the large model $L$, or a held state $\varnothing$. Ownership records which prediction enters the generation state. Query selection determines the computation needed to obtain that prediction, and the executing model's KV buffer supplies context at other positions. Each model maintains its own representation of the shared state, so progress made through one model can condition a subsequent call to the other.

\paragraph{Mask-selection objective.}
\label{subsec:mask-selection}
Selection concentrates expensive prediction work at positions indicated by the feedback, while allowing reliable inexpensive candidates to advance generation. Continuous refinement uses a spatial budget to assign large-model updates alongside small-model predictions. Discrete generation selects candidates for commitment as confidence and block progress evolve. The mask specifies prediction work; model-specific cached context supplies the conditioning needed to execute it.

\subsection{\name{} Design}
\name{} implements a unified token-level model allocation framework to accommodate for different properties of the two diffusion backends. In T2I generation, a hybrid refinement stage ($H$) lies between large-only ($L$) and small-only ($S$) denoising. In dLLMs, model selection is revisited across decoding calls as token confidence and commitment progress evolve.

\subsubsection{Token Routing}
Both tasks route prediction work at token level. Feedback determines the candidate positions and model predictions used to advance the shared state. The continuous policy adopts candidates from both models within a hybrid step; the discrete policy selects a model call and then commits eligible token predictions. In each case, routing specifies local work while cached context supports its execution.

At iteration $s$, the current generation state $x_s$ provides eligible positions $\mathcal U_s$: the image latent grid or the remaining masked positions of an active text block. The router assigns each eligible position $i$ a feedback score:
\begin{equation}
g_s(i)=
\begin{cases}
\operatorname{Mean}_{\mathrm{ch}}|\bar X_s^L(i)-\bar X_s^S(i)|,
 &\text{T2I},\\
\bar c_s(i), &\text{dLLM}
\end{cases}.
\label{eq:routing-score}
\end{equation}
Here, $\bar X_s^L(i)$ and $\bar X_s^S(i)$ are the latest available candidate vectors; $\operatorname{Mean}_{\mathrm{ch}}$ averages absolute channel-wise differences. $\bar c_s(i)$ is the latest prediction confidence. High disagreement and $1-\bar c_s(i)$ both indicate uncertainty, with confidence thresholds selecting discrete candidates. Full evaluations initialize continuous feedback; unscored discrete positions enter mandatory selection. The router reuses this feedback until a model call supplies updated predictions, placing the selection decision before the next evaluation.

For model $m\in\{S,L\}$ (small or large), selected positions $\mathcal A_s^m$ and binary mask $M_s^m$ follow the same interface:
\begin{equation}
\begin{aligned}
\mathcal A_s^m&=
\begin{cases}
\operatorname{TopK}(g_s,k_s^m),
 &\text{T2I},\\
\operatorname{Thresh}(g_s,\tau_q^m)\cup\mathcal F_s,
 &\text{dLLM},
\end{cases}\\
M_s^m(i)&=\mathbb I[i\in\mathcal A_s^m].
\end{aligned}
\label{eqa:mask-decision}
\end{equation}
For hybrid refinement, $k_s^m=\max(1,\lfloor\rho_s^m|\mathcal U_s|\rfloor)$ with fraction $\rho_s^m\in(0,1]$. $\operatorname{TopK}$ returns exactly $k_s^m$ positions with the highest scores; ties follow the operator's index selection. $\operatorname{Thresh}$ selects scored eligible positions reaching query threshold $\tau_q^m$. The mandatory set $\mathcal F_s\subseteq\mathcal U_s$ contains unscored positions and the active commitment frontier. Feedback computed after an update determines the next selection.
\Cref{table:state-realizations} specifies their update rules: continuous latents combine large-model refinement with small-model drafts, while discrete states incorporate candidates that pass the commitment threshold.

\begin{table}[!t]
\centering
\small
\setlength\tabcolsep{3pt}
\caption{Two realizations of token ownership.}
\label{table:state-realizations}
\begin{tabular}{@{}p{.23\linewidth}p{.34\linewidth}p{.34\linewidth}@{}}
\toprule
Dimension & Continuous (T2I) & Discrete (dLLM)\\
\midrule
Feedback & Disagreement & Confidence\\
Selection & Budgeted TopK & Threshold and frontier\\
Owner $\pi_s(i)$ & Mask-selected candidate & Committing model; $\varnothing$ holds\\
Scheduling & Both models per hybrid step & One model per request call\\
Shared state & Merged latent & Committed tokens\\
Context & \multicolumn{2}{p{.70\linewidth}@{}}{Model-specific KV and selected queries}\\
\bottomrule
\end{tabular}
\end{table}

\textbf{Continuous refinement.}
During hybrid execution, the small model processes the full latent and the large model refines the selected high-disagreement positions: $\mathcal A_s^S=\mathcal U_s$, with $\mathcal A_s^L$ selected by the budgeted rule. The hybrid budget fixes the selected fraction; stage transitions change overall model coverage. Following adaptive switching~\cite{cheng2025srdiffusionacceleratevideodiffusion}, stage transitions use the relative change of the clean-state estimate:
\begin{equation}
D_s=\frac{\|\hat x_s^0-\hat x_{s-1}^0\|_2}
{\max(\|\hat x_s^0\|_2,\delta)}.
\label{equ:diff}
\end{equation}
Here, $\hat x_s^0$ is the sampler's clean-state estimate and $\delta>0$ is a numerical floor. Consecutive estimates within a stage determine progress; $D_s<\eta_p$ triggers the next stage. Disagreement scores select spatial refinement work within the hybrid stage.

For example, a region where the candidate updates closely agree can advance with the small-model output, while a neighboring high-disagreement region receives the large-model update. Both candidates start from the same current latent, and their selected outputs form a single next latent. The updated disagreement field selects the mask for the following iteration. A fixed budget can thus follow different regions as their refinement demands change.

\textbf{Discrete commitment.}
Within the dLLM hybrid stage, each request chooses its next model from confidence and commitment progress. After a large call, at least $b$ remaining positions with confidence reaching $\tau_r$ trigger small-model execution. Small-model calls continue when they commit tokens, the maximum remaining selected confidence reaches $\tau_c$, and fewer than $J$ consecutive small calls have occurred; an empty-set maximum is $-\infty$. Otherwise, the large model refines the state. Small-ready requests are batched at a target size, waiting-time bound, or when all active requests are ready, preserving each request's routing decision.

For example, a block with several confident candidates can advance through successive small-model calls. Each commitment removes a position from the remaining masks and adds its token to the conditioning sequence. The next call evaluates remaining candidates in that updated context. A subsequent large-model call supplies predictions used to reconsider the allocation, while all committed IDs remain in the shared block.


\subsubsection{Context-Preserving KV Cache}
Selected queries retain global context through model-specific layer-wise KV buffers (\Cref{fig:overview}b). Let $\mathcal R_s^m$ contain required refresh positions, including uninitialized entries. The executed queries are
\begin{equation}
\mathcal Q_s^m=\Pi(\mathcal A_s^m)\cup\mathcal R_s^m.
\label{eq:dllm-query-set}
\end{equation}
Here, $\Pi$ maps output positions to transformer query tokens: for patchified latents, a patch is selected when it contains a selected latent position; for text, $\Pi$ is the identity. The output mask remains on the latent grid for merging. Selected queries refresh their KV while other positions retain their latest entries. For discrete tokens, refresh also covers IDs changed since that model's last evaluation. Refresh supplies context; selected eligible positions supply generation candidates.

Consider a masked text position $i$ admitted by the query threshold: it receives fresh logits, and the commitment threshold determines whether its prediction enters the sequence. A previously committed position $j$ whose ID changed since this model's last call enters cache refresh. Its representation provides context for predicting $i$. Candidate queries, context refresh, and output commitment therefore serve distinct roles within one evaluation.

At a model handoff, changed-ID refresh uses the receiving model's own input record. A token committed by the other model enters the receiving model's refresh set when its stored ID differs from the shared sequence. The call recomputes that representation alongside selected masked positions. Refreshed committed tokens contribute context, while eligible masked positions supply candidates for commitment. Progress passes through the shared IDs, and each model computes the KV representations used by its own attention layers.

At layer $\ell$, fresh and retained entries form the completed token buffers:
\begin{equation}
\begin{aligned}
&(\widetilde K_{s,\ell}^m(i),\widetilde V_{s,\ell}^m(i))\\
&\quad=
\begin{cases}
(K_{s,\ell}^{m,\mathrm{new}}(i),V_{s,\ell}^{m,\mathrm{new}}(i)),
 &i\in\mathcal Q_s^m,\\
(K_{\ell}^{m,\mathrm{cache}}(i),V_{\ell}^{m,\mathrm{cache}}(i)),
 &i\notin\mathcal Q_s^m
\end{cases}.
\end{aligned}
\label{eq:kv-cache-completion}
\end{equation}
Cache indices denote transformer token positions. Concatenating completed-prefix KV, $K_\ell^{m,\mathrm{ext}},V_\ell^{m,\mathrm{ext}}$, supplies the full allowed context:
\begin{equation}
\begin{aligned}
K_{s,\ell}^{m,\mathrm{ctx}}&=[K_\ell^{m,\mathrm{ext}};\widetilde K_{s,\ell}^m],\\
V_{s,\ell}^{m,\mathrm{ctx}}&=[V_\ell^{m,\mathrm{ext}};\widetilde V_{s,\ell}^m].
\end{aligned}
\label{eq:dllm-context}
\end{equation}
Here, $[\,;\,]$ concatenates context positions. The query matrix $Q_{s,\ell}^m$ contains rows for $\mathcal Q_s^m$ and attends to the completed KV:
\begin{equation}
\begin{aligned}
\operatorname{Attn}_{s,\ell}^m
&=\operatorname{Softmax}(S_{s,\ell}^m)V_{s,\ell}^{m,\mathrm{ctx}},\\
S_{s,\ell}^m
&=\frac{Q_{s,\ell}^m(K_{s,\ell}^{m,\mathrm{ctx}})^\top}{\sqrt d}
+B_{s,\ell}^m.
\end{aligned}
\label{eq:masked-attention}
\end{equation}
The head dimension is $d$, and $B_{s,\ell}^m$ is the backend attention mask, with zero for allowed interactions and $-\infty$ for excluded interactions. Token-wise operators such as FFNs, normalization, and output projections process the active query rows~\cite{distrifusion,mixfusion}, and their predictions are scattered to the corresponding token positions.

Context reuse exploits temporal redundancy~\cite{distrifusion,fang2024xdit,liu2026dllmcache,ma2025dkvcache}. Each model rebuilds completed text blocks from final IDs before appending their KV to the prefix; the runtime batches this maintenance with active queries and prefill.

\subsubsection{Model Execution and State Update}
\label{subsubsec:switch}
\label{subsubsec:dllm-routing}
Let $a_s^m(i)$ be model $m$'s candidate next-state value. Both state types adopt updates through the same ownership map:
\begin{equation}
x_{s+1}(i)=
\begin{cases}
a_s^{\pi_s(i)}(i),&\pi_s(i)\neq\varnothing,\\
x_s(i),&\pi_s(i)=\varnothing.
\end{cases}
\label{eq:ownership-update}
\end{equation}
For images, $a_s^m=X_s^m$ is the sampler's candidate next latent. During hybrid refinement, $\pi_s(i)=L$ when $M_s^L(i)=1$ and $S$ otherwise, giving
\begin{equation}
x_{s+1}=M_s^L\odot X_s^L+(\mathbf 1-M_s^L)\odot X_s^S.
\label{eq:image-state-update}
\end{equation}
Here, $\odot$ denotes token-wise multiplication, broadcasting across channels. Full-large and small-only stages assign all positions to $L$ and $S$, respectively.

For an Euler sampler, selecting candidate next latents is equivalent to selecting their predicted velocities before the update. Both candidates use the same input latent and signed step; their masked sum retains this common base and selects the velocity contribution at each position. The merged latent becomes the input to both models at the following iteration, and the candidate differences provide feedback for the next mask.

For discrete commitment, model logits $o_s^m(i)$ give confidence $c_s^m(i)=\max_v\operatorname{Softmax}(o_s^m(i))_v$, where $v$ indexes the vocabulary. The committed set is $\mathcal C_s^m=\{i\in\mathcal A_s^m:c_s^m(i)>\tau_s^m\}$, with model- and phase-specific commitment threshold $\tau_s^m$. When this set is empty, the large-model progress policy adds $\arg\max_{i\in\mathcal A_s^L}c_s^L(i)$ to $\mathcal C_s^L$. The state update is
\begin{equation}
x_{s+1}(i)=
\begin{cases}
\arg\max_v o_s^m(i)_v,&i\in\mathcal C_s^m,\\
x_s(i),&\text{otherwise}
\end{cases}.
\label{eq:text-state-update}
\end{equation}
Here, $\pi_s(i)=m$ for $i\in\mathcal C_s^m$ and $\varnothing$ otherwise; the owned candidate is $a_s^m(i)=\arg\max_v o_s^m(i)_v$. 

\begin{table*}[!t]
\centering
\small
\setlength\tabcolsep{3pt}
\caption{dLLM end-to-end results.}
\label{table:e2e-dllm}
\begin{tabular}{llrrrrr}
\toprule
Task & Method & Score & Tok/s & Speedup & GF/tok & Mean/p90 (s)\\
\midrule
GSM8K & Plain SDAR & 91.5 & 308 & 1.00$\times$ & 179.1 & 12.1/19.4\\
 & dKV adaptation & 90.0 & 333 & 1.08$\times$ & 127.5 & 11.1/17.3\\
 & SimSD & 88.0 & 320 & 1.04$\times$ & 121.4 & 12.6/18.1\\
 & \name{} & 89.5 & 354 & 1.15$\times$ & 90.2 & 13.7/20.8\\
\midrule
MBPP & Plain SDAR & 54.5 & 188 & 1.00$\times$ & 330.1 & 6.3/9.3\\
 & dKV adaptation & 51.0 & 222 & 1.18$\times$ & 205.1 & 5.3/7.9\\
 & SimSD & 48.5 & 253 & 1.34$\times$ & 172.4 & 6.6/10.4\\
 & \name{} & 56.0 & 263 & 1.40$\times$ & 113.0 & 5.8/9.4\\
\midrule
MATH500 & Plain SDAR & 65.8 & 626 & 1.00$\times$ & 139.3 & 20.3/33.1\\
 & dKV adaptation & 66.0 & 797 & 1.27$\times$ & 102.5 & 16.0/27.1\\
 & SimSD & 59.4 & 774 & 1.24$\times$ & 99.7 & 17.0/28.6\\
 & \name{} & 65.6 & 991 & 1.58$\times$ & 69.4 & 16.8/27.6\\
\midrule
CNN/DM & Plain SDAR & 27.2 & 224 & 1.00$\times$ & 296.5 & 9.0/12.3\\
 & dKV adaptation & 27.3 & 285 & 1.27$\times$ & 196.7 & 7.1/9.9\\
 & SimSD & 26.7 & 305 & 1.36$\times$ & 171.9 & 7.9/12.2\\
 & \name{} & 27.0 & 327 & 1.46$\times$ & 112.4 & 7.4/9.7\\
\bottomrule
\end{tabular}
\end{table*}

\paragraph{Ownership and execution granularity.}
T2I adopts different model candidates within one hybrid step and revisits positions during refinement. dLLM chooses one model per request call; its committed tokens become conditioning positions. The adopted predictions determine the next shared latent or sequence, allowing subsequent calls to incorporate progress made by either model.

%% file: sec/4_eval.tex
\section{Evaluation}

We evaluate generation quality and efficiency, isolate the benefits of allocation and partial execution, and examine how model participation changes across token positions, iterations, and budgets.

\subsection{Experimental Setup}
\label{subsec:setup}
\textbf{T2I.}
We use Stable Diffusion 3.5 Large and SD3 Medium~\cite{esser2024scalingrectifiedflowtransformers}, generating one $768\times768$ image for each of 5,000 sampled captions from COCO and DiffusionDB~\cite{lin2015microsoftcococommonobjects,wangDiffusionDBLargescalePrompt2023}. Runs use 50 denoising steps on an 80\,GB A100. T-Stitch switches from small to large at 40\,\% of the steps~\cite{pan2025tstitch}; SRDiffusion uses a large-to-small threshold of 0.01~\cite{cheng2025srdiffusionacceleratevideodiffusion}. \name{} uses $\rho_s^L=0.3$ during hybrid refinement and stage thresholds $(0.5,0.3)$ for $D_s$ in \Cref{equ:diff}. We report FID against each dataset's original images as the reference distribution, CLIP prompt alignment, LPIPS similarity to large-model outputs~\cite{fid,clipscore,lpips}, and image latency.

\textbf{dLLM.}
We use SDAR-1.7B and SDAR-8B Chat~\cite{cheng2025sdarsynergisticdiffusionautoregressionparadigm} with 32-token blocks on an RTX 6000 Ada GPU and shared fused variable-length kernels. All methods use execution batch capacity 24. Decode GF/token estimates query work as $2(P_LQ_L+P_SQ_S)/G$, with model parameters $P_m$, decode queries $Q_m$, and generated tokens $G$. The cache-only baseline adapts dKV-Cache-Decode to SDAR~\cite{ma2025dkvcache}. Scores are percentages: answer accuracy for GSM8K and MATH500~\cite{cobbe2021gsm8k,math500}, pass@1 for MBPP~\cite{austin2021program}, and ROUGE-L F1 for CNN/DM~\cite{cnn-dm}. GSM8K and MBPP each use 200 prompts, similar as \cite{cui2026simsdsimplespeculativedecoding}. The throughput counting includes the prefill stage, and speedup is relative to plain SDAR at the same batch capacity. Mean/p90 denote request latency in seconds.

\subsection{Quality and Efficiency}
\begin{table}[!t]
\centering
\small
\setlength\tabcolsep{3pt}
\caption{SD3 end-to-end results. Latency is seconds per image.}
\label{table:t2i-e2e}
\begin{tabular}{lrrrr}
\toprule
Method & FID $\downarrow$ & CLIP $\uparrow$ & LPIPS $\downarrow$ & Lat. $\downarrow$\\
\midrule
\multicolumn{5}{c}{COCO}\\
Large & 27.64 & 31.81 & -- & 13.64\\
T-Stitch & 31.87 & 32.01 & 0.70 & 9.63\\
SRDiffusion & 31.67 & 31.89 & 0.73 & 9.43\\
\name{} & 30.43 & 31.84 & 0.42 & 8.11\\
\midrule
\multicolumn{5}{c}{DiffusionDB}\\
Large & 25.72 & 32.93 & -- & 13.64\\
T-Stitch & 29.34 & 32.96 & 0.73 & 9.61\\
SRDiffusion & 26.97 & 32.89 & 0.70 & 9.45\\
\name{} & 25.06 & 32.87 & 0.56 & 8.12\\
\bottomrule
\end{tabular}
\end{table}

\textbf{T2I results.}
\name{} improves the quality--efficiency balance over full-image switching (\Cref{table:t2i-e2e}). Latency is 8.11--8.12\,s per image, giving 1.68$\times$ speedup over the large model and about 14\,\% lower latency than SRDiffusion. On COCO, FID decreases from SRDiffusion's 31.67 to 30.43; on DiffusionDB, it decreases from 26.97 to 25.06. LPIPS is also lower than both switching baselines, while CLIP remains close to the large-model score. Selective refinement retains large-model predictions at chosen positions as the small model advances the remaining latent. 

\textbf{dLLM results.}
Across reasoning, code generation, and summarization, \name{} achieves 1.15--1.58$\times$ plain-SDAR throughput with 50--66\,\% lower estimated decode query cost (\Cref{table:e2e-dllm}). GSM8K and MBPP throughput increases from 308 to 354 and from 188 to 263 tokens/s, respectively. On MATH500, throughput rises from 626 to 991 tokens/s with scores of 65.8 and 65.6; CNN/DM retains similar ROUGE-L with 1.46$\times$ throughput. Against SimSD, throughput is 4--28\,\% higher across all four tasks, with comparable or better quality scores. The scheduler groups small-ready requests to amortize model-call overhead. This throughput-oriented policy explains the request-latency tradeoff in GSM8K, where mean latency is 13.7\,s versus plain SDAR's 12.1\,s: requests may wait while a small-model batch forms.

\begin{figure}[!t]
\centering
\includegraphics[width=\linewidth]{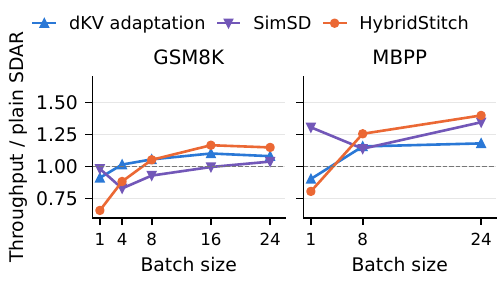}
\caption{GSM8K and MBPP throughput scaling relative to plain SDAR.}
\label{fig:dllm-concurrency}
\end{figure}

\textbf{dLLM batching.}
\Cref{fig:dllm-concurrency} shows how computation savings translate into throughput at different batch sizes. Relative throughput for dKV and \name{} rises with concurrency on GSM8K and MBPP, whereas SimSD's curves first decrease and then recover. We attribute this contrast to the interaction between model work and device utilization: speculative verification exposes parallel work at small batches, while cache reuse and capacity allocation reduce work as more requests share the GPU. At low concurrency, fixed call overhead narrows the execution-time difference between the small and large models, and an additional small-model call contributes appreciable overhead. Larger batches amortize these calls and make the reduced query work more consequential, translating token-level savings into sustained throughput.

\subsection{Component Contributions}
\label{subsec:abalation}
\begin{table}[!t]
\centering
\small
\setlength\tabcolsep{3pt}
\caption{SD3 ablations on COCO. StaticMask freezes routing; FullLarge computes all large-model queries but merges selected outputs.}
\label{table:t2i-ablation}
\begin{tabular}{lrrrr}
\toprule
Method & FID & CLIP & LPIPS & Lat.\\
\midrule
StaticMask & 31.88 & 31.79 & 0.45 & 7.72\\
FullLarge & 31.32 & 31.84 & 0.39 & 8.97\\
\name{} & 30.43 & 31.84 & 0.42 & 8.11\\
\bottomrule
\end{tabular}
\end{table}

\textbf{T2I.}
\Cref{table:t2i-ablation} distinguishes adaptive output selection from selected-query execution. StaticMask fixes the selected region during hybrid refinement. Updating that region improves FID from 31.88 to 30.43 and LPIPS from 0.45 to 0.42, with latency increasing from 7.72 to 8.11\,s. The quality gain supports tracking prediction differences as the latent evolves, even when the selected fraction is fixed.

FullLarge retains mask-based output merging but computes the entire tokens with large model. Executing only the selected queries with cache completion reduces latency from 8.97 to 8.11\,s, while the quality scores remain similar to FullLarge. The latency reduction demonstrates that KV cache effectively lowers computation while preserving quality close to full-model execution.

\textbf{dLLM.}
The dKV adaptation provides cached large-model decoding as the comparison in \Cref{table:e2e-dllm}. \name{} increases throughput over dKV by 6--24\,\% across all four tasks and reduces estimated query cost by a further 29--45\,\%. On MBPP, query cost decreases from 205.1 to 113.0 GF/token as throughput rises from 222 to 263 tokens/s. Small-model commitments advance the shared block directly, reducing the prediction work subsequently assigned to the large model alongside KV reuse.

\subsection{Allocation Behavior}
\begin{figure}[t]
\centering
\includegraphics[width=\linewidth]{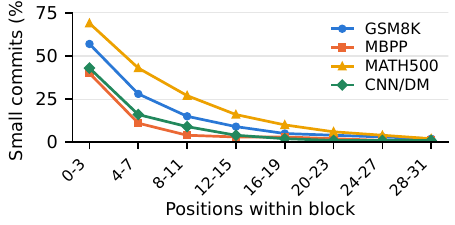}
\caption{Small-model ownership by block position.}
\label{fig:dllm-commits}
\end{figure}

\Cref{fig:dllm-commits} locates small-model updates within a text block. In the instrumented runs, the small model commits 8.1--21.8\,\% of generated tokens across the four tasks. Its share is largest at positions 0--3 and declines toward the block end in every task. MATH500 has the highest initial share, 69\,\%, compared with 57\,\% on GSM8K, 40\,\% on MBPP, and 43\,\% on CNN/DM. Positional percentages use each position group's tokens as the denominator; query cost measures execution work.

\begin{figure}[t]
\centering
{\small\textbf{T2I: stage-boundary distributions}\par}
\includegraphics[width=\linewidth]{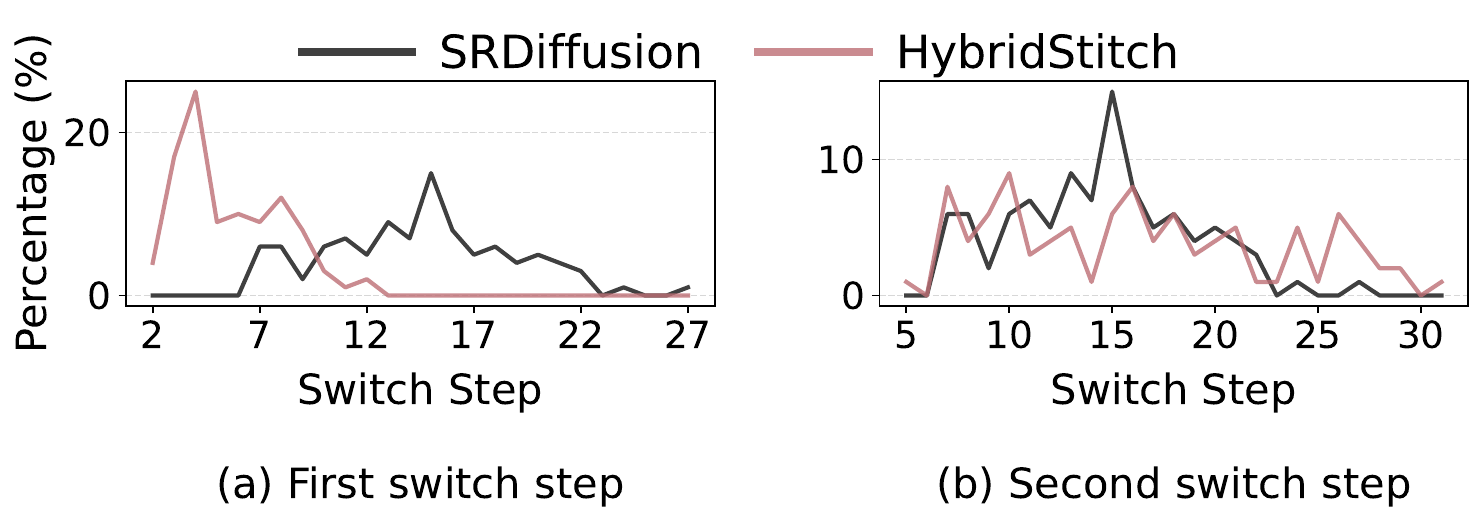}
{\small\textbf{dLLM: token commitments over block iterations}\par}
\includegraphics[width=\linewidth]{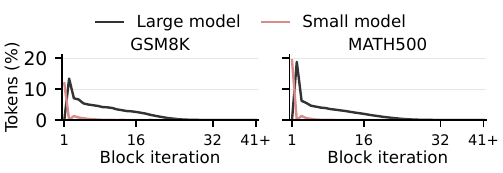}
\caption{T2I ownership-budget transitions (top) and dLLM committed owners by block iteration (bottom).}
\label{fig:diff}
\end{figure}

\textbf{T2I.}
The upper panels of \Cref{fig:diff} show T2I's two stage boundaries. The first introduces budgeted hybrid refinement; the second assigns the full latent to the small model. Hybrid refinement starts before SRDiffusion's full-image transition, so inexpensive updates begin while selected regions continue to receive large-model predictions.

\textbf{dLLM.}
The lower panels of \Cref{fig:diff} measure commitments rather than stage transitions: dLLM generation remains in one hybrid stage. Each curve counts tokens committed by its model at a block iteration, normalized by all generated tokens. For both GSM8K and MATH500, small-model commitments concentrate early, while large-model calls account for updates across later iterations. Confidence and commitment progress therefore change model participation within the block, rather than switching the entire decoder into a small-only phase.

\begin{figure}[!t]
\centering
\includegraphics[width=\linewidth]{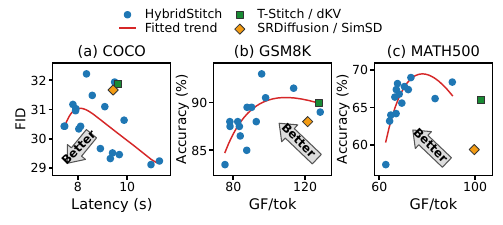}
\caption{Quality--cost sweeps with fitted trends: (a) T2I, (b) GSM8K, and (c) MATH500. }
\label{fig:tradeoff}
\end{figure}

\textbf{Quality--efficiency operating points.}
\Cref{fig:tradeoff} provides an exploratory view of allocation budgets: spatial fractions and stage thresholds for T2I, and commitment thresholds for dLLMs. The curves summarize the trends of \name{}. All panels expose the quality--efficiency tradeoff controlled by ownership allocation. For all models and datasets, \name{} demonstrates better accuracy/efficiency trends compared to baselines. \name{} achieves a larger throughput improvement on MATH500 than on GSM8K, consistent with the higher small-model commitment share observed in Figure 5.

%% file: sec/5_relate.tex

%% file: sec/6_conclusion.tex
\section{Conclusion}
We have presented \name{}, an ownership-based capacity allocation framework for continuous and discrete diffusion. Generation feedback selects the model supplying each adopted update, while model-specific cached context supports partial execution. Adaptive spatial masks and confident token commitments reduce large-model prediction work, and selected-query execution turns that allocation into the measured generation-efficiency gains.

%% file: main.bib
@String(CVPR  = {IEEE Conf. Comput. Vis. Pattern Recog.})

@String(ICCV  = {Int. Conf. Comput. Vis.})

@String(CVPR  = {CVPR})

@String(ICCV  = {ICCV})


%% file: ref/misc.bib
@misc{cache_dit,
  title={cache-dit: A PyTorch-native and Flexible Inference Engine with Hybrid Cache Acceleration and Parallelism for DiTs.},
  url={https://github.com/vipshop/cache-dit.git},
  note={Open-source software available at https://github.com/vipshop/cache-dit.git},
  author={DefTruth, vipshop.com},
  year={2025}
}


%% file: ref/ml.bib
@misc{podell2023sdxlimprovinglatentdiffusion,
      title={SDXL: Improving Latent Diffusion Models for High-Resolution Image Synthesis}, 
      author={Dustin Podell and Zion English and Kyle Lacey and Andreas Blattmann and Tim Dockhorn and Jonas Müller and Joe Penna and Robin Rombach},
      year={2023},
      eprint={2307.01952},
      archivePrefix={arXiv},
      primaryClass={cs.CV},
      url={https://arxiv.org/abs/2307.01952}, 
}

@inproceedings{
liu2026dllmcache,
title={d{LLM}-Cache: Accelerating Diffusion Large Language Models with Adaptive Caching},
author={Zhiyuan Liu and Yicun Yang and Yaojie Zhang and Junjie Chen and Chang Zou and Qingyan Wei and Shaobo Wang and Yichen Zhu and Linfeng Zhang},
booktitle={Forty-third International Conference on Machine Learning},
year={2026},
url={https://openreview.net/forum?id=guLrgDb2jB}
}

@inproceedings{
ma2025dkvcache,
title={d{KV}-Cache: The Cache for Diffusion Language Models},
author={Xinyin Ma and Runpeng Yu and Gongfan Fang and Xinchao Wang},
booktitle={The Thirty-ninth Annual Conference on Neural Information Processing Systems},
year={2025},
url={https://openreview.net/forum?id=Gppo2JImHs}
}

@misc{gao2025selfspeculativedecodingdiffusion,
      title={Self Speculative Decoding for Diffusion Large Language Models}, 
      author={Yifeng Gao and Ziang Ji and Yuxuan Wang and Biqing Qi and Hanlin Xu and Linfeng Zhang},
      year={2025},
      eprint={2510.04147},
      archivePrefix={arXiv},
      primaryClass={cs.CL},
      url={https://arxiv.org/abs/2510.04147}, 
}

@inproceedings{cnn-dm,
  author={Karl Moritz Hermann and Tomás Kociský and Edward Grefenstette and Lasse Espeholt and Will Kay and Mustafa Suleyman and Phil Blunsom},
  title={Teaching Machines to Read and Comprehend},
  year={2015},
  cdate={1420070400000},
  pages={1693-1701},
  url={http://papers.nips.cc/paper/5945-teaching-machines-to-read-and-comprehend},
  booktitle={NIPS},
}

@inproceedings{math500,
 author = {Hendrycks, Dan and Burns, Collin and Kadavath, Saurav and Arora, Akul and Basart, Steven and Tang, Eric and Song, Dawn and Steinhardt, Jacob},
 booktitle = {Proceedings of the Neural Information Processing Systems Track on Datasets and Benchmarks},
 editor = {J. Vanschoren and S. Yeung},
 pages = {},
 title = {Measuring Mathematical Problem Solving With the MATH Dataset},
 url = {https://datasets-benchmarks-proceedings.neurips.cc/paper_files/paper/2021/file/be83ab3ecd0db773eb2dc1b0a17836a1-Paper-round2.pdf},
 volume = {1},
 year = {2021}
}

@article{austin2021program,
  title={Program Synthesis with Large Language Models},
  author={Austin, Jacob and Odena, Augustus and Nye, Maxwell and Bosma, Maarten and Michalewski, Henryk and Dohan, David and Jiang, Ellen and Cai, Carrie and Terry, Michael and Le, Quoc and others},
  journal={arXiv preprint arXiv:2108.07732},
  year={2021}
}

@article{cobbe2021gsm8k,
  title={Training Verifiers to Solve Math Word Problems},
  author={Cobbe, Karl and Kosaraju, Vineet and Bavarian, Mohammad and Chen, Mark and Jun, Heewoo and Kaiser, Lukasz and Plappert, Matthias and Tworek, Jerry and Hilton, Jacob and Nakano, Reiichiro and Hesse, Christopher and Schulman, John},
  journal={arXiv preprint arXiv:2110.14168},
  year={2021}
}

@misc{cui2026simsdsimplespeculativedecoding,
      title={SimSD: Simple Speculative Decoding in Diffusion Language Models}, 
      author={Junxia Cui and Haotian Ye and Runchu Tian and Hongcan Guo and Jinya Jiang and Haoru Li and Chaojie Ren and Yiming Huang and Kaijie Zhu and Zhongkai Yu and Kun Zhou and Jingbo Shang},
      year={2026},
      eprint={2606.02544},
      archivePrefix={arXiv},
      primaryClass={cs.CL},
      url={https://arxiv.org/abs/2606.02544}, 
}

@misc{ye2025dream7bdiffusionlarge,
      title={Dream 7B: Diffusion Large Language Models}, 
      author={Jiacheng Ye and Zhihui Xie and Lin Zheng and Jiahui Gao and Zirui Wu and Xin Jiang and Zhenguo Li and Lingpeng Kong},
      year={2025},
      eprint={2508.15487},
      archivePrefix={arXiv},
      primaryClass={cs.CL},
      url={https://arxiv.org/abs/2508.15487}, 
}

@misc{nie2025largelanguagediffusionmodels,
      title={Large Language Diffusion Models}, 
      author={Shen Nie and Fengqi Zhu and Zebin You and Xiaolu Zhang and Jingyang Ou and Jun Hu and Jun Zhou and Yankai Lin and Ji-Rong Wen and Chongxuan Li},
      year={2025},
      eprint={2502.09992},
      archivePrefix={arXiv},
      primaryClass={cs.CL},
      url={https://arxiv.org/abs/2502.09992}, 
}

@misc{cheng2025sdarsynergisticdiffusionautoregressionparadigm,
      title={SDAR: A Synergistic Diffusion-AutoRegression Paradigm for Scalable Sequence Generation}, 
      author={Shuang Cheng and Yihan Bian and Dawei Liu and Linfeng Zhang and Qian Yao and Zhongbo Tian and Wenhai Wang and Qipeng Guo and Kai Chen and Biqing Qi and Bowen Zhou},
      year={2025},
      eprint={2510.06303},
      archivePrefix={arXiv},
      primaryClass={cs.LG},
      url={https://arxiv.org/abs/2510.06303}, 
}

@misc{labs2025flux1kontextflowmatching,
      title={FLUX.1 Kontext: Flow Matching for In-Context Image Generation and Editing in Latent Space},
      author={Black Forest Labs and Stephen Batifol and Andreas Blattmann and Frederic Boesel and Saksham Consul and Cyril Diagne and Tim Dockhorn and Jack English and Zion English and Patrick Esser and Sumith Kulal and Kyle Lacey and Yam Levi and Cheng Li and Dominik Lorenz and Jonas Müller and Dustin Podell and Robin Rombach and Harry Saini and Axel Sauer and Luke Smith},
      year={2025},
      eprint={2506.15742},
      archivePrefix={arXiv},
      primaryClass={cs.GR},
      url={https://arxiv.org/abs/2506.15742},
}

@misc{flux2024,
    author={Black Forest Labs},
    title={FLUX},
    year={2024},
    howpublished={\url{https://github.com/black-forest-labs/flux}},
}

@misc{wu2025qwenimagetechnicalreport,
      title={Qwen-Image Technical Report}, 
      author={Chenfei Wu and Jiahao Li and Jingren Zhou and Junyang Lin and Kaiyuan Gao and Kun Yan and Sheng-ming Yin and Shuai Bai and Xiao Xu and Yilei Chen and Yuxiang Chen and Zecheng Tang and Zekai Zhang and Zhengyi Wang and An Yang and Bowen Yu and Chen Cheng and Dayiheng Liu and Deqing Li and Hang Zhang and Hao Meng and Hu Wei and Jingyuan Ni and Kai Chen and Kuan Cao and Liang Peng and Lin Qu and Minggang Wu and Peng Wang and Shuting Yu and Tingkun Wen and Wensen Feng and Xiaoxiao Xu and Yi Wang and Yichang Zhang and Yongqiang Zhu and Yujia Wu and Yuxuan Cai and Zenan Liu},
      year={2025},
      eprint={2508.02324},
      archivePrefix={arXiv},
      primaryClass={cs.CV},
      url={https://arxiv.org/abs/2508.02324}, 
}

@misc{esser2024scalingrectifiedflowtransformers,
      title={Scaling Rectified Flow Transformers for High-Resolution Image Synthesis}, 
      author={Patrick Esser and Sumith Kulal and Andreas Blattmann and Rahim Entezari and Jonas Müller and Harry Saini and Yam Levi and Dominik Lorenz and Axel Sauer and Frederic Boesel and Dustin Podell and Tim Dockhorn and Zion English and Kyle Lacey and Alex Goodwin and Yannik Marek and Robin Rombach},
      year={2024},
      eprint={2403.03206},
      archivePrefix={arXiv},
      primaryClass={cs.CV},
      url={https://arxiv.org/abs/2403.03206}, 
}

@article{team2025zimage,
  title={Z-Image: An Efficient Image Generation Foundation Model with Single-Stream Diffusion Transformer},
  author={Z-Image Team},
  journal={arXiv preprint arXiv:2511.22699},
  year={2025}
}

@article{liu2025decoupled,
  title={Decoupled DMD: CFG Augmentation as the Spear, Distribution Matching as the Shield},
  author={Dongyang Liu and Peng Gao and David Liu and Ruoyi Du and Zhen Li and Qilong Wu and Xin Jin and Sihan Cao and Shifeng Zhang and Hongsheng Li and Steven Hoi},
  journal={arXiv preprint arXiv:2511.22677},
  year={2025}
}

@article{jiang2025distribution,
  title={Distribution Matching Distillation Meets Reinforcement Learning},
  author={Jiang, Dengyang and Liu, Dongyang and Wang, Zanyi and Wu, Qilong and Jin, Xin and Liu, David and Li, Zhen and Wang, Mengmeng and Gao, Peng and Yang, Harry},
  journal={arXiv preprint arXiv:2511.13649},
  year={2025}
}

@InProceedings{distrifusion,
    author    = {Li, Muyang and Cai, Tianle and Cao, Jiaxin and Zhang, Qinsheng and Cai, Han and Bai, Junjie and Jia, Yangqing and Li, Kai and Han, Song},
    title     = {DistriFusion: Distributed Parallel Inference for High-Resolution Diffusion Models},
    booktitle = {Proceedings of the IEEE/CVF Conference on Computer Vision and Pattern Recognition (CVPR)},
    month     = {June},
    year      = {2024},
    pages     = {7183-7193}
}

@article{fang2024xdit,
  title={xDiT: an Inference Engine for Diffusion Transformers (DiTs) with Massive Parallelism},
  author={Fang, Jiarui and Pan, Jinzhe and Sun, Xibo and Li, Aoyu and Wang, Jiannan},
  journal={arXiv preprint arXiv:2411.01738},
  year={2024}
}

@inproceedings{
ma2024learningtocache,
title={Learning-to-Cache: Accelerating Diffusion Transformer via Layer Caching},
author={Xinyin Ma and Gongfan Fang and Michael Bi Mi and Xinchao Wang},
booktitle={The Thirty-eighth Annual Conference on Neural Information Processing Systems},
year={2024},
url={https://openreview.net/forum?id=ZupoMzMNrO}
}

@InProceedings{TaylorSeer,
    author    = {Liu, Jiacheng and Zou, Chang and Lyu, Yuanhuiyi and Chen, Junjie and Zhang, Linfeng},
    title     = {From Reusing to Forecasting: Accelerating Diffusion Models with TaylorSeers},
    booktitle = {Proceedings of the IEEE/CVF International Conference on Computer Vision (ICCV)},
    month     = {October},
    year      = {2025},
    pages     = {15853-15863}
}

@inproceedings{
pan2025tstitch,
title={T-Stitch: Accelerating Sampling in Pre-Trained Diffusion Models with Trajectory Stitching},
author={Zizheng Pan and Bohan Zhuang and De-An Huang and Weili Nie and Zhiding Yu and Chaowei Xiao and Jianfei Cai and Anima Anandkumar},
booktitle={The Thirteenth International Conference on Learning Representations},
year={2025},
url={https://openreview.net/forum?id=2mqb8bPHeb}
}

@misc{cheng2025srdiffusionacceleratevideodiffusion,
      title={SRDiffusion: Accelerate Video Diffusion Inference via Sketching-Rendering Cooperation}, 
      author={Shenggan Cheng and Yuanxin Wei and Lansong Diao and Yong Liu and Bujiao Chen and Lianghua Huang and Yu Liu and Wenyuan Yu and Jiangsu Du and Wei Lin and Yang You},
      year={2025},
      eprint={2505.19151},
      archivePrefix={arXiv},
      primaryClass={cs.GR},
      url={https://arxiv.org/abs/2505.19151}, 
}

@misc{lin2015microsoftcococommonobjects,
      title={Microsoft COCO: Common Objects in Context}, 
      author={Tsung-Yi Lin and Michael Maire and Serge Belongie and Lubomir Bourdev and Ross Girshick and James Hays and Pietro Perona and Deva Ramanan and C. Lawrence Zitnick and Piotr Dollár},
      year={2015},
      eprint={1405.0312},
      archivePrefix={arXiv},
      primaryClass={cs.CV},
      url={https://arxiv.org/abs/1405.0312}, 
}

@inproceedings{fid,
 author = {Heusel, Martin and Ramsauer, Hubert and Unterthiner, Thomas and Nessler, Bernhard and Hochreiter, Sepp},
 booktitle = {Advances in Neural Information Processing Systems},
 editor = {I. Guyon and U. Von Luxburg and S. Bengio and H. Wallach and R. Fergus and S. Vishwanathan and R. Garnett},
 pages = {},
 publisher = {Curran Associates, Inc.},
 title = {GANs Trained by a Two Time-Scale Update Rule Converge to a Local Nash Equilibrium},
 url = {https://proceedings.neurips.cc/paper_files/paper/2017/file/8a1d694707eb0fefe65871369074926d-Paper.pdf},
 volume = {30},
 year = {2017}
}

@inproceedings{clipscore,
    title = "{CLIPS}core: A Reference-free Evaluation Metric for Image Captioning",
    author = "Hessel, Jack  and
      Holtzman, Ari  and
      Forbes, Maxwell  and
      Le Bras, Ronan  and
      Choi, Yejin",
    editor = "Moens, Marie-Francine  and
      Huang, Xuanjing  and
      Specia, Lucia  and
      Yih, Scott Wen-tau",
    booktitle = "Proceedings of the 2021 Conference on Empirical Methods in Natural Language Processing",
    month = nov,
    year = "2021",
    address = "Online and Punta Cana, Dominican Republic",
    publisher = "Association for Computational Linguistics",
    url = "https://aclanthology.org/2021.emnlp-main.595/",
    doi = "10.18653/v1/2021.emnlp-main.595",
    pages = "7514--7528"
}

@article{zhangefficient,
  title={Efficient Attention Methods: Hardware-efficient, Sparse, Compact, and Linear Attention},
  author={Zhang, Jintao and Su, Rundong and Liu, Chunyu and Wei, Jia and Wang, Ziteng and Wang, Haoxu and Zhang, Pengle and Jiang, Huiqiang and Huang, Haofeng and Xiang, Chendong and others},
  year={2025}
}

@article{zhang2026sla2,
  title={SLA2: Sparse-Linear Attention with Learnable Routing and QAT},
  author={Zhang, Jintao and Wang, Haoxu and Jiang, Kai and Zheng, Kaiwen and Jiang, Youhe and Stoica, Ion and Chen, Jianfei and Zhu, Jun and Gonzalez, Joseph E},
  journal={arXiv preprint arXiv:2602.12675},
  year={2026}
}

@INPROCEEDINGS{lpips,
  author={Zhang, Richard and Isola, Phillip and Efros, Alexei A. and Shechtman, Eli and Wang, Oliver},
  booktitle={2018 IEEE/CVF Conference on Computer Vision and Pattern Recognition}, 
  title={The Unreasonable Effectiveness of Deep Features as a Perceptual Metric}, 
  year={2018},
  volume={},
  number={},
  pages={586-595},
  doi={10.1109/CVPR.2018.00068}}

@inproceedings{
zhang2026sla,
title={{SLA}: Beyond Sparsity in Diffusion Transformers via Fine-Tunable Sparse{\textendash}Linear Attention},
author={Jintao Zhang and Haoxu Wang and Kai Jiang and Shuo Yang and Kaiwen Zheng and Haocheng Xi and Ziteng Wang and Hongzhou Zhu and Min Zhao and Ion Stoica and Joseph E. Gonzalez and Jun Zhu and Jianfei Chen},
booktitle={The Fourteenth International Conference on Learning Representations},
year={2026},
url={https://openreview.net/forum?id=eD8IPvNoZB}
}

@inproceedings{
zhang2025spargeattn,
title={SpargeAttn: Training-Free Sparse Attention Accelerating Any Model Inference},
author={Jintao Zhang and Chendong Xiang and Haofeng Huang and Jia wei and Haocheng Xi and Jun Zhu and Jianfei Chen},
booktitle={Sparsity in LLMs (SLLM): Deep Dive into Mixture of Experts, Quantization, Hardware, and Inference},
year={2025},
url={https://openreview.net/forum?id=UZggtUfsJV}
}

@inproceedings{
xi2025sparse,
title={Sparse Video-Gen: Accelerating Video Diffusion Transformers with Spatial-Temporal Sparsity},
author={Haocheng Xi and Shuo Yang and Yilong Zhao and Chenfeng Xu and Muyang Li and Xiuyu Li and Yujun Lin and Han Cai and Jintao Zhang and Dacheng Li and Jianfei Chen and Ion Stoica and Kurt Keutzer and Song Han},
booktitle={Forty-second International Conference on Machine Learning},
year={2025},
url={https://openreview.net/forum?id=u8CA3qIS0V}
}

@inproceedings{wangDiffusionDBLargescalePrompt2023,
  title = {{{DiffusionDB}}: {{A}} Large-Scale Prompt Gallery Dataset for Text-to-Image Generative Models},
  booktitle = {Proceedings of the 61st Annual Meeting of the Association for Computational Linguistics (Volume 1: {{Long}} Papers)},
  author = {Wang, Zijie J. and Montoya, Evan and Munechika, David and Yang, Haoyang and Hoover, Benjamin and Chau, Duen Horng},
  year = {2023},
  url = {https://aclanthology.org/2023.acl-long.51}
}


%% file: ref/sys.bib
@inproceedings{mixfusion,
author = {Sun, Desen and Zhao, Zepeng and Wang, Yuke},
title = {MixFusion: A Patch-Level Parallel Serving System for Mixed-Resolution Diffusion Models},
year = {2026},
isbn = {9798400723100},
publisher = {Association for Computing Machinery},
address = {New York, NY, USA},
url = {https://doi.org/10.1145/3774934.3786420},
doi = {10.1145/3774934.3786420},
booktitle = {Proceedings of the 31st ACM SIGPLAN Annual Symposium on Principles and Practice of Parallel Programming},
pages = {522–536},
numpages = {15},
location = {Sydney, NSW, Australia},
series = {PPoPP '26}
}

@misc{sun2024flexcacheflexibleapproximatecache,
      title={FlexCache: Flexible Approximate Cache System for Video Diffusion}, 
      author={Desen Sun and Henry Tian and Tim Lu and Sihang Liu},
      eprint={2501.04012},
      archivePrefix={arXiv},
      primaryClass={cs.MM},
      url={https://arxiv.org/abs/2501.04012}, 
      journal={arXiv preprint arXiv:2501.04012},
      year={2024}
}

@inproceedings {nirvana,
author = {Shubham Agarwal and Subrata Mitra and Sarthak Chakraborty and Srikrishna Karanam and Koyel Mukherjee and Shiv Kumar Saini},
title = {Approximate Caching for Efficiently Serving {Text-to-Image} Diffusion Models},
booktitle = {21st USENIX Symposium on Networked Systems Design and Implementation (NSDI 24)},
year = {2024},
isbn = {978-1-939133-39-7},
address = {Santa Clara, CA},
pages = {1173--1189},
url = {https://www.usenix.org/conference/nsdi24/presentation/agarwal-shubham},
publisher = {USENIX Association},
month = apr
}

@inproceedings{modm,
author = {Xia, Yuchen and Sharma, Divyam and Yuan, Yichao and Kundu, Souvik and Talati, Nishil},
title = {MoDM: Efficient Serving for Image Generation via Mixture-of-Diffusion Models},
year = {2025},
isbn = {9798400721656},
publisher = {Association for Computing Machinery},
address = {New York, NY, USA},
url = {https://doi.org/10.1145/3760250.3762220},
doi = {10.1145/3760250.3762220},
booktitle = {Proceedings of the 31st ACM International Conference on Architectural Support for Programming Languages and Operating Systems, Volume 1},
pages = {163–182},
numpages = {20},
location = {USA},
series = {ASPLOS '26}
}
